\documentclass[11pt]{article}
\usepackage{coling2018}
\usepackage{times}
\usepackage{url}
\usepackage{hyperref}
\usepackage{latexsym}
\usepackage{graphicx}
\usepackage{float} 
\usepackage{subfigure}

\title{Sogou Machine Reading Comprehension Toolkit}

\author{Jindou Wu, Yunlun Yang, Chao Deng, Hongyi Tang \\
 {\bf Bingning Wang},{\bf Haoze Sun},{\bf Ting Yao},{\bf Qi Zhang} \\ Sogou, Inc. \\
{\tt \{wujindou,yangyunlun,dengchao,tanghongyi\}} \\
 {\tt \{wangbingning,haozesun,yaoting,qizhang\}@sogou-inc.com}
 }

\date{}

\begin{document}
\maketitle
\begin{abstract}
Machine reading comprehension have been intensively studied in recent years, and neural network-based models have shown dominant performances. 
In this paper, we present a Sogou Machine Reading Comprehension (SMRC) toolkit that can be used to provide the fast and efficient development of modern machine comprehension models, including both published models and original prototypes.
To achieve this goal, the toolkit provides dataset readers, a flexible preprocessing pipeline, necessary neural network components, and built-in models, which make the whole process of data preparation, model construction, and training easier.
\end{abstract}

\section{Introduction}
Given a passage (context) and a question about it, a reading comprehension system should be able to read the passage and answer the question. While not a hard task for a human, it requires that the system both understand natural language and have knowledge about the world. 
Because of the renaissance of neural networks and accessibility of large-scale datasets, great progress has recently been made in reading comprehension. For example, according to the leaderboard of SQuAD 1.0 \cite{DBLP:conf/emnlp/RajpurkarZLL16}, over $80$ systems have been submitted, and human performance has been left behind.
In experiments, the reimplementation and comparison of these solutions are necessary but not easy tasks, because researchers usually build their blocks from scratch and in different environments. Meanwhile, the efficient construction of original prototypes is not possible, although reading comprehension models often share similar components and architectures.

In this paper, we present the Sogou Machine Reading Comprehension toolkit\footnote{\url{https://github.com/sogou/SMRCToolkit}}, which has the goal of allowing the rapid and efficient development of modern machine comprehension models, including both published models and original prototypes.
First, the toolkit simplifies the dataset reading process by providing dataset reader modules that support popular datasets. Second, the flexible preprocessing pipeline allows vocabulary building, linguistic feature extraction, and operations to work in a seamless way.
Third, the toolkit offers frequently used neural network components, a trainer module, and a save/load function, which accelerates the construction of custom models. Last, but not the least, some published models are implemented in the toolkit, making model comparison and modification convenient. The toolkit is built based on the Tensorflow\footnote{\url{https://github.com/tensorflow/tensorflow}} library \cite{DBLP:conf/osdi/AbadiBCCDDDGIIK16}.

\begin{figure*}[htbp] 
\centering 
\includegraphics[width=\textwidth]{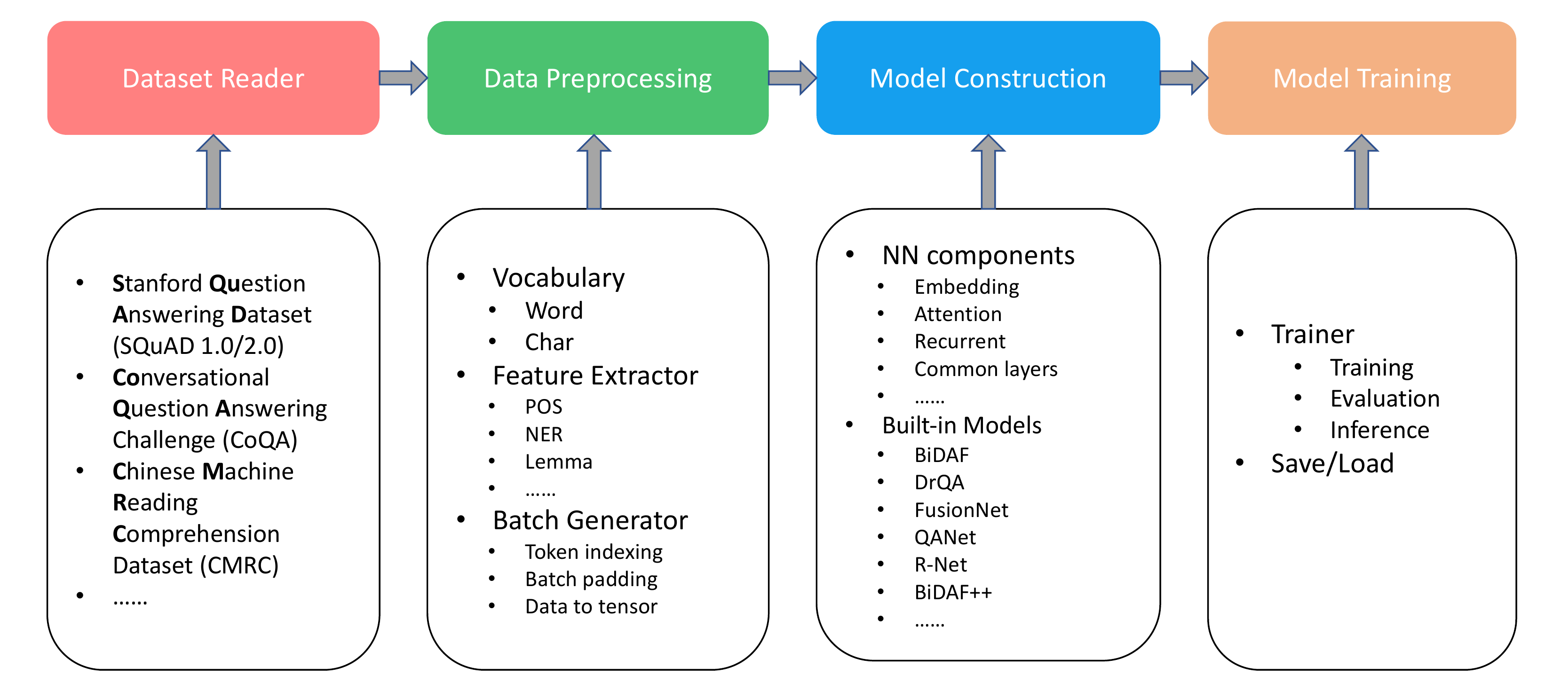}
\caption{Toolkit Architecture} 
\label{architecture} 
\end{figure*}

\section{Toolkit Framework}
As shown in Figure 1, the architecture of our toolkit mainly contains four modules: the
Dataset Reader, Data Preprocessing, Model Construction, and Model Training modules.  These four modules
are designed as a pipeline flow and can be used for most machine reading comprehension tasks. In the following, we
will introduce each part in detail.
\subsection{Dataset Reader}
One reason that machine reading comprehension has made rapid progress, that cannot be ignored, is the release of a variety of large-scale and high-quality question answering datasets. In addition, preprocessing and evaluating are essential steps when doing research on these datasets.

\noindent \textbf{Reader} \indent To avoid repeating the development of dataset reading codes, the toolkit provides reader modules for some typical datasets SQuAD 1.0 \cite{DBLP:conf/emnlp/RajpurkarZLL16}, SQuAD 2.0 \cite{DBLP:conf/acl/RajpurkarJL18} and CoQA \cite{DBLP:journals/corr/abs-1808-07042}. To enhance the language diversity, we also support a Chinese dataset, CMRC2018 \cite{DBLP:journals/corr/abs-1810-07366}.
The reader modules first tokenize texts and generate labels (e.g., start/end positions), and then transform data instances into nested structure objects, the fields of which are uniformly named. This makes data serialization/deserialization convenient and helps in error analysis. By inheriting the base reader, users can develop custom readers for other datasets.

\noindent \textbf{Evaluator} \indent Most datasets offer official evaluation scripts. To ease the model validation and early stopping, we integrate these evaluation scripts into the toolkit and simplify the evaluation in the model training process.

\subsection{Data Preprocessing}
To prepare data for the training model, we need to build a vocabulary, extract linguistic features, and map discrete features into indices. The toolkit provides modules for fulfilling these requirements.

\noindent \textbf{Vocabulary Builder} \indent By scanning the training data from the dataset reader, Vocabulary Builder maintains a corresponding vocabulary of words (and characters if needed). Adding any special tokens or setting the whole vocabulary is allowed as well. Another important function of Vocabulary Builder is creating an embedding matrix from pretrained word embeddings. If you feed a pretrained embedding file, for example Glove\footnote{\url{https://nlp.stanford.edu/projects/glove/}} \cite{DBLP:conf/emnlp/PenningtonSM14}, to the Vocabulary Builder, it will produce a word embedding matrix for its inner vocabulary.

\noindent \textbf{Feature Extractor} \indent Linguistic features are used in many machine reading comprehension models such as DrQA \cite{DBLP:conf/acl/ChenFWB17}, FusionNet \cite{DBLP:journals/corr/abs-1711-07341} and have been proven to be effective. The Feature Extractor supports commonly used features, e.g., part-of-speech (POS) tags, called entity recognition (NER) tags, along with normalized term frequency (TF) and word-level exact matching. By simply adding new feature fields, Feature Extractor does not break the serializability and readability of data instance objects. Meanwhile, Feature Extractor also builds vocabularies for discrete features like POS and NER, which will be used in the next steps for index mapping.

\noindent \textbf{Batch Generator} \indent The last step of the preprocessing is to pack all of the features up and modify them to fit the form of the model input. In Batch Generator, we first map words and tags to indices, pad length-variable features, transform all of the features into tensors, and then batch them. To make these steps efficient, we implement Batch Generator based on the Tensorflow Dataset API\footnote{\url{https://www.tensorflow.org/api_docs/python/tf/data/Dataset}}, which parallelizes data transformation and provides fundamental functions like dynamic padding and data shuffling, which make it behave consistently with the “generator” in Python. Batch Generator is designed to be flexible and compatible with the feature types frequently used in machine reading comprehension tasks.

\subsection{Model Construction}
The core part of the machine reading comprehension task is constructing an effective and efficient model for generating answers from given passages. The toolkit provides two methods: build your own model or use a built-in model. For the first one, we implement frequently used neural network components in the machine reading comprehension task. We follow the idea of functional API and wrap them as MRC specific supplements of Tensorflow layers.

\noindent \textbf{Embedding} \indent Besides a vanilla embedding layer, the toolkit also provides \textit{PartiallyTrainableEmbedding}, as used in \cite{DBLP:conf/acl/ChenFWB17} \cite{DBLP:journals/corr/abs-1711-07341}, and pretrained contextualized representation layers, including \textit{CoVeEmbedding}, \textit{ElmoEmbedding}, and \textit{BertEmbedding}.

\noindent \textbf{Recurrent} \indent \textit{BiLSTM} and \textit{BiGRU} are basic recurrent layers, and their CuDNN version \textit{CudnnBiLSTM} and \textit{CudnnBiGRU} are also available.

\noindent \textbf{Similarity Function} \indent Functions are available for calculating the word-level similarities between texts, e.g., \textit{DotProduct}, \textit{TriLinear}, and \textit{MLP}.

\noindent \textbf{Attention} \indent Attention layers are usually used together with the Similarity Function, e.g., \textit{BiAttention}, \textit{UniAttention}, and \textit{SelfAttention}.

\noindent \textbf{Basic Layer} \indent Some of the basic layers are used in machine reading comprehension models, e.g., \textit{VariationalDropout}, and \textit{Highway}, \textit{ReduceSequence}.

\noindent \textbf{Basic Operation} \indent These are mainly masking operations, e.g., \textit{masked\_softmax}, \textit{mask\_logits}. By inheriting the base model class and combining the components above, developers should be able to construct most mainstream machine reading comprehension models. To build a custom model, a developer should define the following three member methods,

\begin{enumerate}
  \item \textit{\_build\_graph}: Define the forward process of the model
  \item \textit{compile}: Schedule the optimization of the model like the learning rate decay and gradient clipping
  \item \textit{get\_best\_answer}: Transform the model output (probability) to answer text
\end{enumerate}

\noindent Training functions (\textit{train\_and\_evaluate}, \textit{evaluate}, and \textit{inference}) should also be inherited if needed.

The toolkit also provides simple interfaces for using the built-in models. We will introduce the details in Section \ref{built-in}.
\subsection{Model Training}
When training a model, we usually care about how the metrics change on the train/dev set, when to perform early stopping, how many epochs the model needs to converge, and so on. Because most models share a similar training strategy, the toolkit provides a Trainer module, with main functions that include baby-sitting the training, evaluation and inference processing, saving the best weights, cooperating with the exponential moving average, and recording the training summary. Each model also provides interfaces for saving and loading the model weights.
\section{Using Built-In Models}\label{built-in}

\subsection{Have a Try}

We will show an example of running the BiDAF model on the SQuAD 1.0 dataset in this section.

First, the data file of SQuAD 1.0 is loaded using SquadReader. Meanwhile, we also create an evaluator for validation.
\begin{figure}[H] 
\centering 
\includegraphics[width=\textwidth]{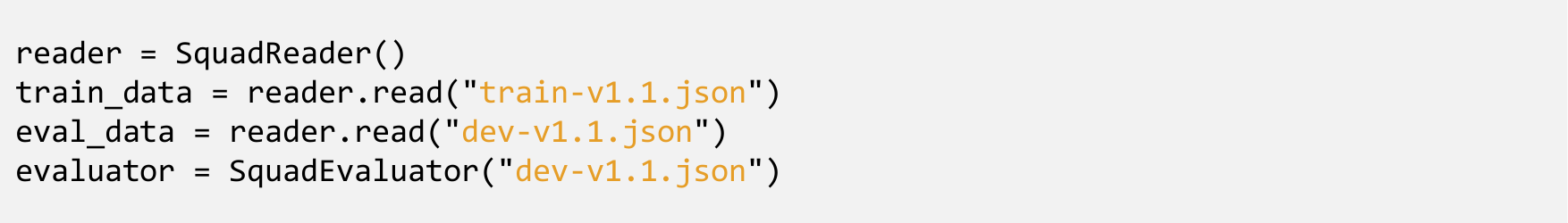}
\end{figure}


Second, we build a vocabulary and corresponding word embedding matrix.
\begin{figure}[H] 
\centering 
\includegraphics[width=\textwidth]{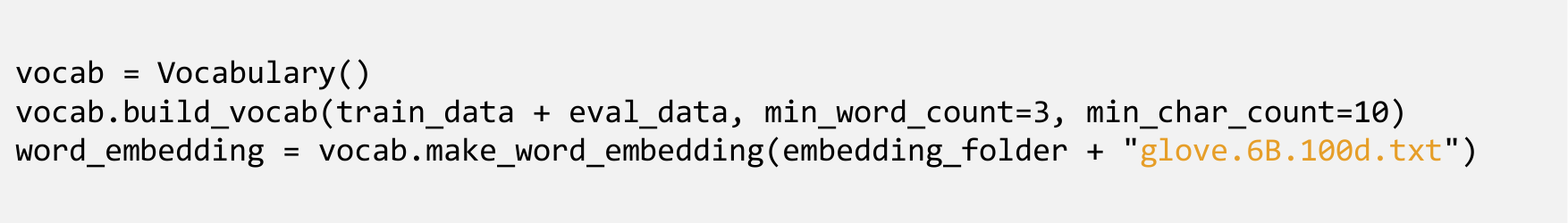}
\end{figure}

Third, data instances are fed to Batch Generator for the necessary preprocessing and batching.

\begin{figure}[H] 
\centering 
\includegraphics[width=\textwidth]{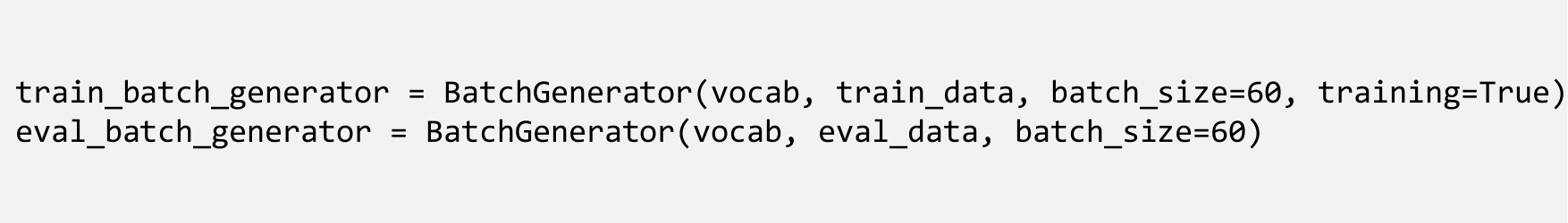}
\end{figure}

Last, we use the built-in BiDAF model and compile it with default hyperparameters. \textit{train\_and\_evaluate} will handle the training process and save the best model weights for inference.
\begin{figure*}[!htbp] 
\centering 
\includegraphics[width=\textwidth]{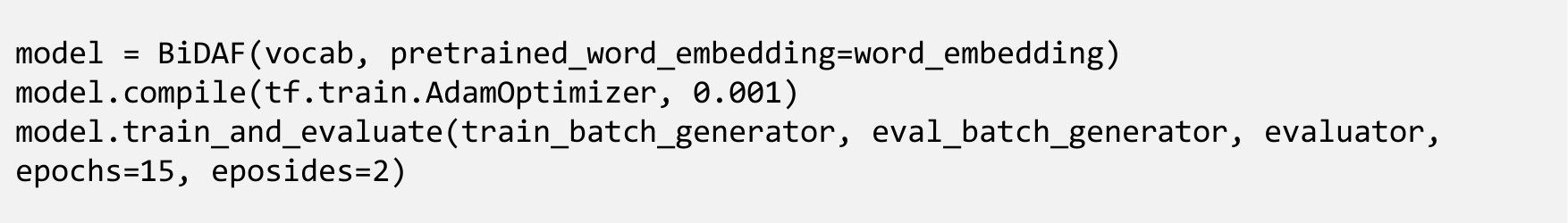}
\end{figure*}

With our toolkit, users can try different machine reading comprehension models in a neat and fast way.

\subsection{Model Zoo}
In the section, we will briefly introduce the machine reading comprehension models implemented in this toolkit. 

\noindent \textbf{BiDAF} was introduced by \cite{DBLP:journals/corr/SeoKFH16}. Unlike the attention mechanisms in previous work, the core idea of BiDAF is bidirectional attention, which models both the query-to-context and context-to-query attention.

\noindent \textbf{DrQA} was proposed by \cite{DBLP:conf/acl/ChenFWB17} and aims at tackling open-domain question answering. DrQA use word embedding, basic linguistic features, and a simple attention mechanism, and proves that simple models without sophisticated architectural designs can also achieve strong results in machine reading comprehension.

\noindent \textbf{FusionNet} Based on an analysis of the attention approaches in previous work, \cite{DBLP:journals/corr/abs-1711-07341} proposed FusionNet, which extends the attention from three perspectives. They proposed the use of the “history of word” and fully aware attention, which let the model combine the information flows from different semantic levels. In addition, the idea was also applied to natural language inference.

\noindent \textbf{R-Net} The main contribution of R-Net was the self-matching attention mechanism. After the gating matching of the context and question, passage self-matching was introduced to aggregate evidence from the whole passage and refine the passage representation.

\noindent \textbf{QANet} The architecture of QANet \cite{DBLP:journals/corr/abs-1804-09541} was adapted from the Transformer \cite{DBLP:conf/nips/VaswaniSPUJGKP17} and only contains the convolution and self-attention. By not using the recurrent Layers, QANet gains a 3–13-fold speed increase in the training time and 4–9-fold increase for the inference time.

\noindent \textbf{IARNN} In our toolkit, two types of Inner Attention-based RNNs (IARNNs) \cite{DBLP:conf/acl/WangL016} are implemented, which are advantageous for sentence representation and efficient in the answer selection task. IARNN-word weights the word representation of the context for the question before inputting into the RNN models. Unlike IARNN-word, which only achieves input word embedding, IARNN-hidden can capture the relationships between multiple words by adding additional context information to the calculation of the attention weights.

\noindent \textbf{BiDAF++} \cite{DBLP:conf/acl/GardnerC18} originally introduced a model for multi-paragraph machine reading comprehension. Based on BiDAF, BiDAF++ adds a self attention layer to increase the model capacity. We also apply the model to CoQA \cite{DBLP:journals/corr/abs-1809-10735} for conversational question answering.

\noindent \textbf{BERT} Pretrained models like BERT \cite{DBLP:journals/corr/abs-1810-04805} and ELMo\cite{DBLP:conf/naacl/PetersNIGCLZ18} have shown great efficacy in many natural language processing tasks. In our toolkit, we use BERT, ELMo, and Cove\cite{DBLP:conf/nips/McCannBXS17} as embedding layers to provide a strong contextualized representation. Meanwhile, we also include the BERT model for machine reading comprehension, as well as our modified version. The results of the models in our toolkit are listed in Section \ref{experiments}.
\section{Experiments}\label{experiments}

We conducted experiments on a supported dataset with the models in the toolkit. By following the experimental settings in the original papers, we attempted to reproduce the results of the models on a different dataset. It is worth mentioning that slight modifications were applied when necessary, and the scripts and hyperparameters for producing the results shown below are included in the toolkit.
\begin{table}[!htbp]
\centering
\caption{F1/EM score on SQuAD 1.0 dev set}\label{tab:squad1}
\begin{tabular}{ | l | c | c |}
	\hline
	Model & toolkit implementation & original paper\\ \hline
	BiDAF & 77.3/67.7  & 77.3/67.7 \\ \hline 
	BiDAF+ELMo & 81.0/72.1 & - \\ \hline
	IARNN-Word & 73.9/65.2 & - \\ \hline
	IARNN-hidden &  72.2/64.3& - \\ \hline 
	DrQA & 78.9/69.4 & 78.8/69.5  \\ \hline 
  DrQA+ELMO&83.1/74.4 & - \\ \hline
	R-Net & 80.0/71.6 & 79.5/71.1  \\ \hline 
	BiDAF++ & 78.6/69.2 & -  \\ \hline 
	FusionNet & 81.0/72.0 & 82.5/74.1  \\ \hline 
	QANet & 80.8/71.8 & 82.7/73.6  \\ \hline 
	BERT-Base & 88.3/80.6 & 88.5/80.8 \\ \hline

\end{tabular}
\end{table}

In Table \ref{tab:squad1}, we report the results of the implemented models on the development set of SQuAD 1.0. A sophisticated and effective attention mechanism is necessary for building a high-performance model, according to the table. In addition,  pretrained models like ELMo and BERT give reading comprehension a big boost and have become a new trend in natural language processing. Our toolkit also wraps commonly used attention and pretrained models in a high-level layer and allows flexible combinations.

\begin{table}[!htbp]
\centering
\caption{F1/EM score on SQuAD 2.0 dev set}\label{tab:squad2}
\begin{tabular}{ | l | c | c |}
	\hline
	Model & toolkit implementation & original paper\\ \hline
	BiDAF & 62.7/59.7 & 62.6/59.8 \\ \hline 
	BiDAF++ & 64.3/61.8 & 64.8/61.9  \\ \hline 
	BiDAF++ + ELMo  & 67.6/64.8& 67.6/65.1 \\ \hline
	BERT-Base & 75.9/73.0 & 75.1/72.0 \\ \hline
\end{tabular}

\end{table}
\begin{table}[!htbp]
\centering
\caption{F1 score on CoQA dev set}\label{tab:coqa}
\begin{tabular}{| l | c | c |}
	\hline
	Model & toolkit implementation & original paper\\ \hline
	BiDAF++ & 71.7 & 69.2 \\ \hline 
	BiDAF++ + ELMo & 74.5 & 69.2 \\ \hline 
	BERT-Base & 78.6 & - \\ \hline
	BERT-Base+Answer Verification & 79.5 & - \\ \hline
\end{tabular}
\end{table}

Because SQuAD 2.0 and CoQA are different from SQuAD 1.0 in a variety of respects, the models are not directly transferrable between these datasets. Following \cite{DBLP:conf/conll/LevySCZ17} and \cite{DBLP:journals/corr/abs-1809-10735}, we implement several effective models.
Moreover, our implemented BiDAF achieves Exact Match $35.0$ and F1 $57.01$ on the CMRC dataset, providing a strong baseline.

\begin{table}[!htbp]
\centering
\caption{F1/EM score of different embedding}\label{tab:embedding}
\begin{tabular}{ | c | c | c |}
	\hline
	Embedding & BiDAF & DrQA \\ \hline
	random  & 70.3/59.3 & 72.4/62.5 \\ \hline
	word2vec & 77.1/67.7 & 78.2/68.8 \\ \hline
	glove   & 77.3/67.7 & 78.9/69.4 \\ \hline
	fast-text-wiki & 77.1/67.7 & 75.4/66.0 \\ \hline
	fast-text-crawl & 76.9/67.4 & 77.0/67.2 \\ \hline
	ELMo    & 79.9/71.1 & 82.7/74.3 \\ \hline 
\end{tabular}
\end{table}

To investigate the effect of the word representation, we selected two popular models and tested their performances with different embeddings. Table \ref{tab:embedding} suggests that DrQA is more sensitive to the word embedding and ELMo helps improve the score consistently (here, when ELMo was used, no word embedding was concatenated).

\section{Conclusion and Future Work}
In the paper, we present the Sogou Machine Reading Comprehension toolkit, which has the goal of allowing the rapid and efficient development of modern machine comprehension models, including both published models and original prototypes.

In the future, we plan to extend the toolkit,  and make it applicable to more tasks, e.g. multi-paragraph and multi-document question answering, and provide more available models.
\bibliographystyle{acl}
\bibliography{coling2018}

\end{document}